\documentclass{article}
\usepackage[nonatbib, final]{nips_2018}
\usepackage[utf8]{inputenc} 
\usepackage[T1]{fontenc} 
\usepackage{hyperref} 
\usepackage{url} 
\usepackage{booktabs} 
\usepackage{amsfonts}      
\usepackage{nicefrac}     
\usepackage{microtype}   
\usepackage[pdftex]{graphicx}
\usepackage{booktabs, array}
\usepackage{amsmath, amsthm, amssymb, latexsym}
\usepackage{xcolor}  
\usepackage{colortbl}  
\usepackage[]{caption}
\DeclareCaptionType[name={Table},listname={}]{supptable}

\DeclareCaptionType[name={Figure},listname={}]{suppfigure}

\usepackage{filecontents}

\title{Integrating omics and MRI data with kernel-based tests and CNNs to identify rare genetic markers for Alzheimer's disease}

\author{
  Stefan~Konigorski \qquad Shahryar~Khorasani \qquad Christoph~Lippert  \\
  Digital Health \& Machine Learning Group, Hasso Plattner Institute \& Universit\"{a}t Potsdam  \\
  Statistical Genomics Group, Max Delbrück Center for Molecular Medicine, Berlin \\ 
  \texttt{\{stefan.konigorski, shahryar.khorasani, christoph.lippert\}@hpi.de}\\
}

\begin{document}

\maketitle

\begin{abstract}
  For precision medicine and personalized treatment, we need to identify predictive markers of disease.
  We focus on Alzheimer's disease (AD), where magnetic resonance imaging scans provide information about the disease status.
  By combining imaging with genome sequencing, we aim at identifying rare genetic markers associated with quantitative traits predicted from convolutional neural networks (CNNs), which traditionally have been derived manually by experts.  Kernel-based tests are a powerful tool for associating sets of genetic variants, but how to optimally model rare genetic variants is still an open research question. We propose a generalized set of kernels that incorporate prior information from various annotations and multi-omics data. In the analysis of data from the Alzheimer’s Disease Neuroimaging Initiative (ADNI), we evaluate whether (i) CNNs yield precise and reliable brain traits, and (ii) the novel kernel-based tests can help to identify loci associated with AD. The results indicate that CNNs provide a fast, scalable and precise tool to derive quantitative AD traits and that new kernels integrating domain knowledge can yield higher power in association tests of very rare variants.
\end{abstract}

\section{Introduction} \label{ref_s1}

In this study, we focus on Alzheimer’s disease (AD) as outcome of interest, which is a progressive neurodegenerative disease, appears late-onset and sporadic in most cases, and is the main cause of dementia in the elderly. As the cognitive symptoms emerge years after the appearance of brain atrophy and exhibit close correlation with the structural changes, brain magnetic resonance imaging (MRI) scans provide a direct way to obtain informative quantitative traits, and fast automated approaches are necessary for large-scale studies. AD has a high estimated heritability of 74\% \cite{gatz1997} and a prevalence of 4.4\% in Europe \cite{lobo2000}. However, the biological pathways underlying AD have not been well-understood and there is yet no known cure. Hence, the identification of AD markers for early detection and as targets for treatment is important. 

For the detection of causal genetic loci, recent sequencing efforts allow in-depth analyses of rare variants in large cohorts, and kernel-based gene-level tests have been proposed for the analysis \cite{ wu2011rare, lee2012optimal, listgarten2013powerful, lippert2014greater, urrutia2015}. They derive similarity scores between samples in the form of a kernel matrix which is computed on a particular genomic locus or functional unit in the genome. Then, kernel-based variance-component test statistics are derived that yield robust and powerful tests. Kernel functions provide a highly flexible way to model genetic variation. However, their full capabilities have not been leveraged and existing approaches still provide suboptimal performance for the analysis of sequencing data \cite{konigorski2017comparison}, where the overwhelming majority of genetic variation is extremely rare. Hence extensions to the existing methods are warranted that leverage the full power of kernels to aggregate the signal of very rare single nucleotide variants (SNVs).

Our contributions in this paper are in two areas. First, we use a convolutional neural network (CNN) to derive quantitative traits from MRI scans in the Alzheimer’s Disease Neuroimaging Initiative (ADNI) and evaluate and compare the obtained traits to traits obtained by the popular yet computationally expensive FreeSurfer software \cite{fischl2002}.
Second, we propose novel kernels for association tests of rare genetic variants that incorporate prior biological knowledge from annotations and multi-omics measures. We perform association analyses between these novel kernels computed on sequencing data and CNN-derived traits as well as other traits to identify genetic loci associated with AD.

\subsection{Related work} \label{ref_s1b}

The association of a set of $m$ genetic markers with a quantitative trait $Y$ with $n$ observations can be tested in a linear mixed model of the form
\begin{equation} 
	Y = X\alpha + Z\beta + G\gamma + \varepsilon, \label{eq1}
\end{equation} 
where $X$ is a covariate design matrix with fixed effects $\alpha$, $\beta \sim \mathcal{N}(0,\sigma_z^2 \,\mathcal{I})$ are random effects of the SNVs in design matrix $Z$
accounting for population stratification, $\mathcal{I}$ is the identity matrix, $\gamma \sim \mathcal{N}(0,\sigma_g^2 \,\mathcal{I})$ are random effects of the $m$ SNVs of interest in the $n \times m$ design matrix $G$, and $\varepsilon \sim \mathcal{N}(0,\sigma_{\varepsilon}^2)$ are error terms. Hence 
\begin{equation} 
	Y \sim \mathcal{N}(X\alpha, \;\sigma_z^2 ZZ^T + \sigma_g^2 K + \sigma_{\varepsilon}^2 \,\mathcal{I} ) \label{eq2} 
\end{equation}
where the $n \times n$ kernel matrix $K=GG^T$ describes the similarity between individuals based on the $m$ SNVs of interest. The association of the $m$ SNVs 
(i.e. $H_0: \sigma_g^2 = 0$ vs. $H_0: \sigma_g^2 > 0$) can be tested using score or likelihood ratio tests. Binary or count traits can be analyzed similarly. 

Popular kernel-based tests include FaST-LMM-Set \cite{listgarten2013powerful, lippert2014greater}, the sequence kernel association test (SKAT, \cite{wu2011rare}) and optimal SKAT (SKAT-O, \cite{lee2012optimal}), which are based on weighted linear kernels $K=GWW^TG^T$ \cite{wu2011rare, listgarten2013powerful, lippert2014greater}, or a linear combination of weighted linear and collapsing kernels \cite{lee2012optimal}. Newer approaches \cite{urrutia2015} derive further data-adaptive combinations of linear, quadratic, IBS, and collapsing kernels. However, all these kernels provide suboptimal performance for the analysis of very rare genetic variants. For example, linear kernels yield uninformative similarity measures (i.e., diagonal kernel matrices for singletons, which are variants with only one observed copy of the minor allele) and collapsing kernels often yield unspecific signals and aggregate noise. 

\section{New kernel-based tests for very rare genetic variants} \label{ref_s2c}

\begin{figure}
	\centering
	\includegraphics[width=1\linewidth]{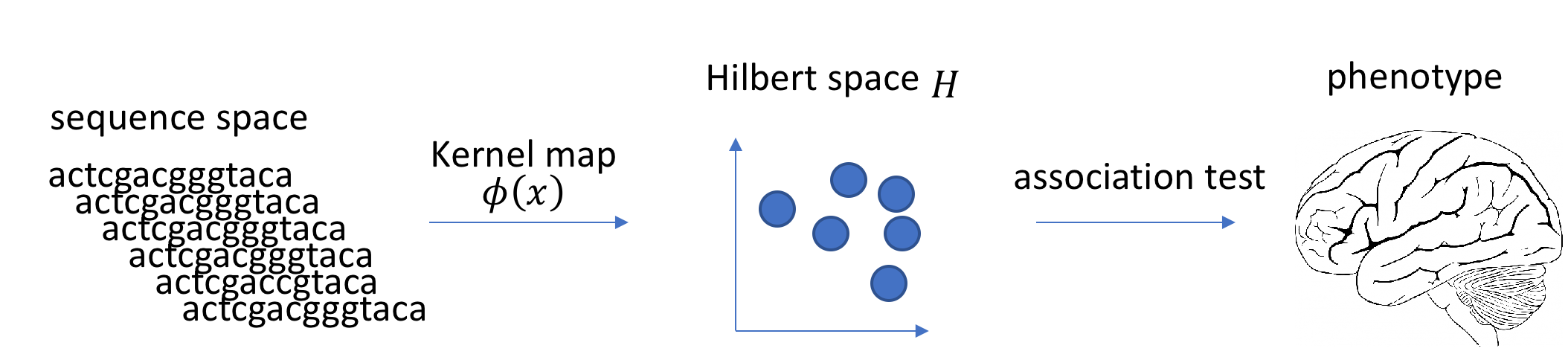}
	\caption{Illustration of association tests using kernel maps.}
	\label{fig1}
\end{figure}

To leverage the full power of kernels computing similarities in high-dimensional Hilbert space, whereto genetic variants are mapped through a potentially infinite-dimensional basis function $\phi$, we consider the more general linear mixed model
\begin{equation} 
Y_i = X_i\alpha + Z_i\beta + \phi(G_i)\gamma + \varepsilon_i, \quad i = 1 \dots n \label{eq3}. 
\end{equation}
Here, $\beta$ and $\gamma$ are normally distributed random effects. After integrating out $\beta$, $\gamma$ and $\epsilon$, it follows that $Y$ is normally distributed with covariance $\sigma_z^2 ZZ^T + \sigma^2_g K + \sigma_{\varepsilon}^2\mathcal{I}$, where we have defined the kernel matrix $K:=\phi(G)\phi(G)^T$. In this model, established tests~\cite{lee2012optimal, listgarten2013powerful, lippert2014greater} can be used to test the association between sets of SNVs and the phenotype, see Figure \ref{fig1} for an illustration.

\subsection{Examples of new kernels}

Let $G$ be the matrix of the $m$ SNVs of interest. We define a class of $n \times n$ kernel matrices $K$ as 
\begin{equation} 
	K= GVWW^TV^TG^T \label{eq4} 
\end{equation}
where different instances are obtained by setting the $m \times m$ weight and similarity matrices $W$, $V$ to the identity, to the matrices outlined below, or any combination of these. See Appendix \ref{Suppl1} for details.

\paragraph{Incorporate annotations}

Set $W$ as the diagonal matrix $W = diag(\sqrt{w_1}, \dots, \sqrt{w_m})$ where $w_j \geq 0$, $j = 1, \dots, m$, is the weight of the $j$-th SNV based on the minor allele frequency (MAF), genomic position, or functional annotation from PolyPhen2 \cite{adzhubei2010}, RegulomeDB \cite{boyle2012}, or others. Set the elements ${ij}$ of $V$ to (i) describe the similarity of SNVs $i$ and $j$ in terms of genomic closeness, or (ii) indicate whether SNVs $i$ and $j$ have a (or the same) functional annotation.

\paragraph{Incorporate information from available omics data}

Set $W = diag(\sqrt{w_1}, \dots, \sqrt{w_m})$ where $w_j \geq 0$, $j = 1, \dots, m$, (i) is the -log$_{10}$ p-value of the association test of the $j$-th SNV with omics data, e.g. gene expression levels of the respective gene (cis-eQTL), or (ii) indicates for each of these p-values if they are $< \alpha$, where $\alpha$ is pre-specified constant, e.g. $\alpha = 0.05$. Set the elements ${ij}$ of $V$ to be indicators whether SNVs $i$ and $j$ both have p-value $<\alpha$.

\section{Application: analysis of ADNI study} \label{ref_s3}

\begin{figure}[!b]
	\centering
	\includegraphics[width=1\linewidth]{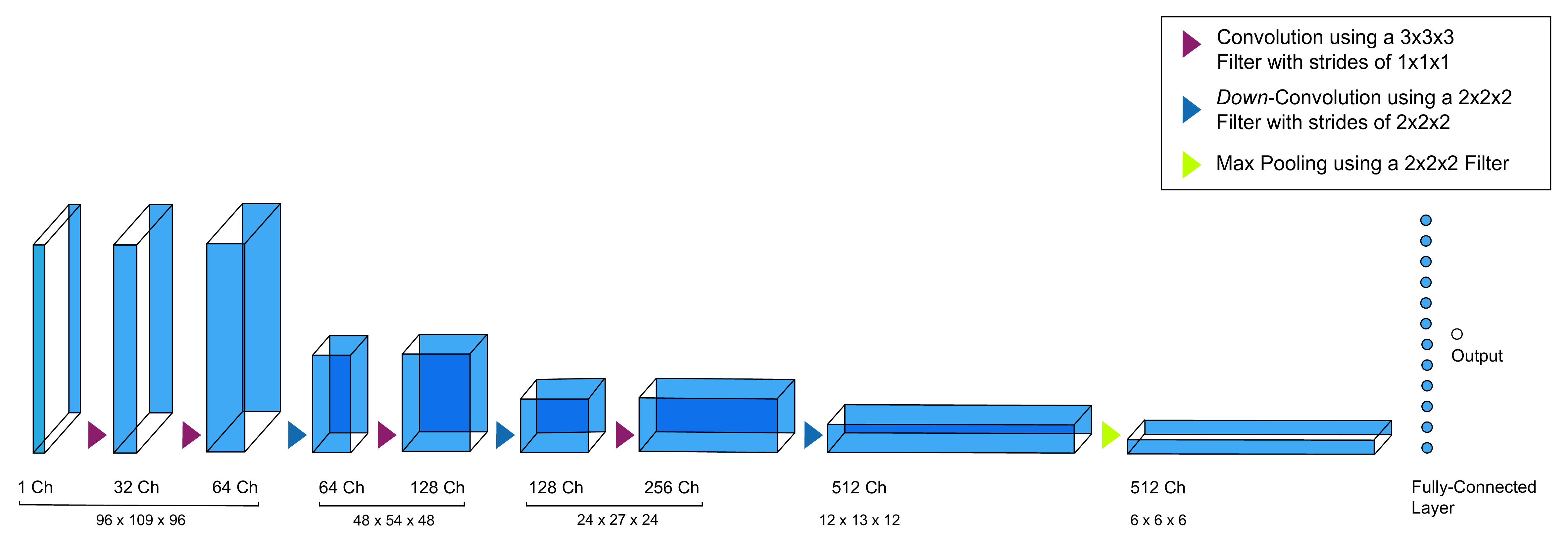}
	\caption{Overview of 3D convolutional neural network.}
	\label{fig2}
\end{figure}

In the application, we analyzed whole-genome-sequencing data, gene expression measures, MRI data as well as AD biomarkers in $n=556$ participants from ADNI, which is a longitudinal study to detect biomarkers and risk factors for AD \cite{weiner2010, weiner2012}.

In a first step, we designed a 3-dimensional CNN comprising seven convolutional layers followed by a max pooling layer and a final fully-connected layer to predict the volume of the 3$^{\text{rd}}$ ventricle from the MRI scans (see Figure \ref{fig2} for an illustration and see Appendix \ref{Suppl2}, Figure \ref{figKeras} for details). To evaluate the approach, we chose the 3$^{\text{rd}}$ ventricle, as we found that the ventricular regions were displayed with a higher contrast and presumably easier to identify. The CNN predicted volume was then used as a quantitative trait in the following genetic association analyses, and evaluated against the predictions by the FreeSurfer software. Both models where trained on a dual Intel Xeon 6148 workstation equipped with an NVidia Titan-V graphics card.

In the main genetic association analysis we analyzed 17,013 (quality-controlled, biallelic, missingness $<$5\%, of any MAF) SNVs in 125 genes in the 1Mbp region around the APOE gene on chromosome 19, similar to the study in \cite{nho2017}, to investigate rare variants in a genomic region where several common variants have been associated with AD. We performed cross-sectional association tests of these 125 genes with 9 different AD traits (peptides CSF A$\beta$, t-tau, p-tau, and the provided brain volumes of entorhinal cortex, hippocampus, medial temporal lobe, ventricles, 3$^{\text{rd}}$ ventricle predicted by FreeSurfer, and 3$^{\text{rd}}$ ventricle predicted from the CNN) adjusting for the covariates age, gender, education, ethnicity, and APOE4 allele. The association tests were performed based on kernels with different combinations of $V$ and $W$ in Appendix \ref{Suppl1} and using standard SKAT and SKAT-O.

\subsection{Results} \label{ref_s3b}

The 125 genes contained on average 220 SNVs ($\min=1$, $\max=1759$).
Of the 17,013 SNVs, 7575 were singletons, 1740 doubletons, and 12,337 SNVs had MAF $<0.01$. 24 participants had dementia, 338 mild cognitive impairment, 194 were cognitive normal (see Table \ref{table2} for descriptive statistics).

In an evaluation of the predicted volume of the 3$^{\text{rd}}$ ventricle, CNN and FreeSurfer predictions showed a high correlation (Pearson $r=0.92$, see Figure \ref{fig3}). For small/large volumes, compared to FreeSurfer, CNN slightly over-/underestimated the volume, which we expect to disappear with larger training data. On the other hand, CNN was much faster (1 second versus 16 hours per scan).

In the main genetic association analyses, a first comparison showed that analyses using the CNN-predicted trait as outcome generally yielded similar and often smaller p-values compared to the FreeSurfer-predicted trait (Figures \ref{fig4}-\ref{fig5}). Preliminary comparisons of the new kernels indicated that the three kernels reported in Table \ref{Table_kernelresults} yielded often the smallest p-values in gene-based tests, hence they are reported here. Tests based on the new kernel 1 yielded consistently smaller or similar p-values for the top genes compared to SKAT and SKAT-O for 8 out of 9 traits (Table \ref{Table_kernelresults}). More detailed comparisons (Figure \ref{fig6}) indicated that while often the same genes were identified with smallest p-value by tests based on the new kernel 1 and by SKAT or SKAT-O, the new kernel 1 also yielded different candidate genes that would not have been identified by SKAT or SKAT-O (and vice versa). The new kernels 2 and 3 yielded sometimes larger but also sometimes much smaller p-values. 

Using a Bonferroni correction (for the 125 tests) of the p-values of the new kernel-based tests, we identified 3 candidate genes for AD with adjusted p-values 0.007, 0.05, 0.07: \texttt{PVR} for CSF t-tau, \texttt{SIX5} for entorhinal cortex and \texttt{PVRL2} for hippocampus.

\begin{table}
	\centering
	\caption{Minimum p-values from the 125 association tests (of the 125 genes) for each respective trait and test. Tests are based on SKAT, SKAT-O and the new kernels 1 (identity $V$, MAF + omics $W$), 2 (genomic distance + omics $V$, omics $W$), 3 (PolyPhen2 + omics $V$, PolyPhen2 + omics + MAF $W$), testing each trait and gene separately. For each trait (row), the smallest p-value is indicated in red.}
	\label{Table_kernelresults}
	\resizebox{0.9\textwidth}{!}{%
		\begin{tabular}{@{}lccccc@{}}
			\toprule

			\textbf{Trait} & 
			\textbf{SKAT-O} & 
			\textbf{SKAT} & 
			\textbf{New Kernel 1} & 
			\textbf{New Kernel 2} & 
			\textbf{New Kernel 3} \\ \midrule 
			
		\cellcolor{gray!10}CSF t-tau &					
		\cellcolor{gray!10}9.1 $\times 10^{-5}$	\qquad & 
		\cellcolor{gray!10}5.8 $\times 10^{-5}$ \qquad & 
		\cellcolor{gray!10}\textcolor{red}{5.2 $\times 10^{-5}$} \qquad & 
		\cellcolor{gray!10}6.4 $\times 10^{-4}$ \qquad & 
		\cellcolor{gray!10}4.0 $\times 10^{-4}$ \qquad \\
			
			CSF p-tau & 					
			1.5 $\times 10^{-3}$ \qquad & 
			9.3 $\times 10^{-4}$ \qquad & 
			\textcolor{red}{8.5 $\times 10^{-4}$} \qquad & 
			1.2 $\times 10^{-3}$ \qquad & 
			1.9 $\times 10^{-3}$ \qquad \\
			
		\cellcolor{gray!10}CSF A$\beta$ &			
		\cellcolor{gray!10}4.9 $\times 10^{-3}$	\qquad & 
		\cellcolor{gray!10}9.9 $\times 10^{-3}$ \qquad & 
		\cellcolor{gray!10}4.9 $\times 10^{-3}$ \qquad & 
		\cellcolor{gray!10}2.1 $\times 10^{-3}$ \qquad & 
		\cellcolor{gray!10}\textcolor{red}{1.7 $\times 10^{-3}$} \qquad \\
			
			Entorhinal cortex &	
			7.0 $\times 10^{-4}$ \qquad & 
			\textcolor{red}{3.3 $\times 10^{-4}$} \qquad & 
			4.2 $\times 10^{-4}$ \qquad & 
			1.6 $\times 10^{-2}$ \qquad & 
			3.4 $\times 10^{-3}$ \qquad \\
			
			\cellcolor{gray!10}Hippocampus &	
			\cellcolor{gray!10}6.6 $\times 10^{-2}$ \qquad & 
			\cellcolor{gray!10}3.7 $\times 10^{-2}$ \qquad & 
			\cellcolor{gray!10}3.8 $\times 10^{-2}$ \qquad & 
			\cellcolor{gray!10}\textcolor{red}{5.6 $\times 10^{-4}$} \qquad & 
			\cellcolor{gray!10}2.5 $\times 10^{-2}$ \qquad \\
			
		Med-temporal lobe&	
		\textcolor{red}{1.1 $\times 10^{-3}$}	\qquad & 
		3.7 $\times 10^{-3}$ \qquad & 
		1.4 $\times 10^{-3}$ \qquad & 
		2.3 $\times 10^{-2}$ \qquad & 
		4.6 $\times 10^{-3}$ \qquad \\
			
			\cellcolor{gray!10}Ventricles 		&	
			\cellcolor{gray!10}1.5 $\times 10^{-2}$ \qquad & 
			\cellcolor{gray!10}9.3 $\times 10^{-3}$ \qquad & 
			\cellcolor{gray!10}1.0 $\times 10^{-2}$ \qquad & 
			\cellcolor{gray!10}\textcolor{red}{9.6 $\times 10^{-4}$} \qquad & 
			\cellcolor{gray!10}5.5 $\times 10^{-3}$ \qquad \\

			FreeS 3$^{\text{rd}}$ Ventricle	&	
			6.8 $\times 10^{-2}$  \qquad 	& 
			6.5 $\times 10^{-2}$  \qquad 	& 
			8.0 $\times 10^{-2}$ \qquad 	& 
			3.4 $\times 10^{-2}$ \qquad 	& 
			\textcolor{red}{5.1 $\times 10^{-3}$} \qquad 	\\

			\cellcolor{gray!10}CNN 3$^{\text{rd}}$ Ventricle	&	
			\cellcolor{gray!10}1.9 $\times 10^{-2}$  \qquad  & 
			\cellcolor{gray!10}5.9 $\times 10^{-2}$  \qquad  & 
			\cellcolor{gray!10}5.9 $\times 10^{-2}$  \qquad & 
			\cellcolor{gray!10}\textcolor{red}{1.7 $\times 10^{-2}$}   \qquad 	& 
			\cellcolor{gray!10}2.0 $\times 10^{-2}$   \qquad 	\\ \bottomrule

		\end{tabular}%
}
\end{table}

\section{Discussion} \label{ref_4}

The empirical analyses indicated that (i) CNNs provide a precise, fast and scalable tool to derive quantitative traits from MRI scans and that (ii) new kernels integrating domain knowledge and omics data constitute a promising approach for the analysis of very rare variants. There is previous evidence for the association of the identified genes with AD \cite{Marioni2018, kwok2018, Hao2018, Beecham2014} to support our findings, and of note, the p-values are much smaller using the new kernels here compared to regular kernels \cite{nho2017}. Limitations of the current analyses are that only few functional annotations are available for rare SNVs, and that only a basic control for population stratification was used. In the interpretation of the results regarding their biological relevance, it can be noted that the analyses were adjusted for the risk factor APOE4, so that the identified genes and SNVs represent markers with independent effects on AD. Future research can investigate kernels measuring the similarity between the bivariate allelic sequences directly and data-adaptive optimal combinations of different kernels. 

\newpage

\bibliographystyle{unsrt}
\bibliography{bibliographyML4H}
\addcontentsline{toc}{section}{References}

\newpage

\section*{Acknowledgments}
\addcontentsline{toc}{section}{Acknowledgments}

Data used in the preparation of this article were obtained from the Alzheimer’s Disease Neuroimaging Initiative (ADNI) database \url{adni.loni.usc.edu}. As such, the investigators within the ADNI contributed to the design and implementation of ADNI and/or provided data but did not participate in analysis or writing of this report. A complete listing of ADNI investigators can be found at: \url{http://adni.loni.usc.edu/wp-content/ uploads/how\_to\_apply/ADNI\_Acknowledgement\_List.pdf}. Data collection and sharing of ADNI was funded by the Alzheimer’s Disease Neuroimaging Initiative (ADNI) (National Institutes of Health Grant U01 AG024904) and DOD ADNI (Department of Defense award number W81XWH-12-2-0012). ADNI is funded by the National Institute on Aging, the National Institute of Biomedical Imaging and Bioengineering, and through generous contributions from the following: Alzheimer’s Association; Alzheimer’s Drug Discovery Foundation; BioClinica Inc; Biogen Idec Inc; Bristol-Myers Squibb Company; Eisai Inc; Elan Pharmaceuticals Inc; Eli Lilly and Company; F. Hoffmann-La Roche Ltd and its affiliated company Genentech Inc; GE Healthcare; Innogenetics N.V.; IXICO Ltd; Janssen Alzheimer Immunotherapy Research \& Development LLC; Johnson \& Johnson Pharmaceutical Research \& Development LLC; Medpace Inc; Merck \& Co Inc; Meso Scale Diagnostics LLC; NeuroRx Research; Novartis Pharmaceuticals Corporation; Pfizer Inc; Piramal Imaging; Servier; Synarc Inc; and Takeda Pharmaceutical Company. The Canadian Institutes of Health Research is providing funds to support ADNI clinical sites in Canada. Private sector contributions are facilitated by the Foundation for the National Institutes of Health (http://www.fnih.org). The grantee organization is the Northern California Institute for Research and Education, and the study is coordinated by the Alzheimer’s Disease Cooperative Study at the University of California, San Diego. ADNI data are disseminated by the Laboratory for Neuro Imaging at the University of Southern California. Samples from the National Cell
Repository for AD (NCRAD), which receives government support under a cooperative agreement grant (U24 AG21886) awarded by the National Institute on Aging (AIG), were used in this study. Funding for the WGS was provided by the Alzheimer’s Association and the Brin Wojcicki Foundation.

\newpage

\appendix
\section*{Supplementary material}
\addcontentsline{toc}{section}{Supplementary material}
\renewcommand{\thesubsection}{\Alph{subsection}}

\subsection{Details on new kernels} \label{Suppl1}

Let $n$ be the number of observations and $m$ the number of SNVs of interest. Define the kernel $K$ as $K= GVWW^TV^TG^T$ in equation (\ref{eq4})
by setting the $m \times m$ matrices $W$ and $V$ to $\mathcal{I}$ or the following.

Consider the weight matrices $W = diag(\sqrt{w_1}, \dots, \sqrt{w_m})$ where

\begin{enumerate}
	\item $w_j \sim Beta(\text{MAF}_j, 1, 25)$ where $Beta$ is the probability density function of the beta distribution and $\text{MAF}_j$ is the minor allele frequency of SNV $j$
	\item $w_j$ is an indicator whether SNV $j$ has a functional annotation in e.g. the PolyPhen2 database
	\item $w_j$ is a numeric encoding of the functional annotation of SNV $j$, for example, in the PolyPhen2 database:
			\begin{equation*} 
			w_j = \left\{ 
				\begin{array}{ll}
				0 & \text{if SNV}_j \text{ is not annotated}\\
				1  & \text{if SNV}_j \text{ has annotation "benign"}\\
				2  & \text{if SNV}_j \text{ has annotation "possibly damaging"}\\
				3  &\text{if SNV}_j \text{ has annotation "probably damaging"}
			\end{array} \right.
			\end{equation*}	
	\item $w_j$ is the -log$_{10}$ p-value from a hypothesis test of the association between SNV $j$ and a variable $Z$ providing relevant information about its biological function, e.g. where $Z$ is the gene expression of the gene in which the SNV lies
	\item $w_j$ is the sum of 1 and an indicator variable whether SNV $j$ is associated with a variable $Z$ providing relevant information about its biological function as in the bullet point above, e.g. evaluating whether the p-value from a hypothesis test of the association between SNV $j$ with a variable $Z$ is smaller than 0.05
	\item $w_j$ is the product of the $w_j$ in bullet points (1 and 4) or (1 and 5)
	\item $w_j$ is the sum of the $w_j$ in bullet points (2 or 3) and (4 or 5)
	\item $w_j$ is the sum of the $w_j$ in bullet points (2 or 3) and 6
\end{enumerate}
with $j = 1, \dots, m$. \\ \\
Consider the $m \times m$ matrices $V$ describing the similarity of SNVs where

\begin{enumerate}
	\item $V_{ij} = $ similarity of SNVs $i$ and $j$ in terms of genomic closeness:
			\begin{equation*} 
				V_{ij} = \left\{ 
					\begin{array}{ll}
						1 & \text{if } i = j\\
						1/d_{ij} & \text{else}
					\end{array} \right.
			\end{equation*}
			where $d_{ij}$ is the genomic distance between SNVs $i$ and $j$ in base pairs
	\item $V_{ij} = $ indicator whether SNVs $i$ and $j$ both have a functional annotation:
			\begin{equation*} 
				V_{ij} = \left\{ 
					\begin{array}{ll}
						1 & \text{if } i = j\\
						1 & \text{if SNVs } i, j \text{ are annotated}\\
						0 & \text{else}.
					\end{array} \right.
			\end{equation*}
	\item $V_{ij} = $ indicator whether SNVs $i$ and $j$ have the same functional annotation
			\begin{equation*} 
				V_{ij} = \left\{ 
					\begin{array}{ll}
						1 & \text{if } i = j\\
						1 & \text{if SNVs } i, j \text{ have the same annotation}\\
						0 & \text{else}.
					\end{array} \right.
			\end{equation*}
	\item $V_{ij} = $ indicator whether SNVs $i$ and $j$ have p-value $<$ specified cutoff value $\alpha$
			\begin{equation*} 
				V_{ij} = \left\{ 
					\begin{array}{ll}
						1 & \text{if } i = j\\
						1 & \text{if SNVs } i, j \text{ both have p-value } < \alpha\\
						0 & \text{else}.
					\end{array} \right.
			\end{equation*}
			where the p-values of SNVs $i, j$ are from association tests with a variable $Z$ that provides relevant information about their biological function, e.g. where $Z$ is the gene expression of the gene in which the SNVs lie
	\item $V$ is the product of the matrices in bullet points (1 and (2 or 3)), (1 and 4), or (4 and (2 or 3))
	\item $V$ is the product of the matrices in bullet points 1 and (2 or 3) and 4		
\end{enumerate}

\subsection{Details on convolutional neural networks} \label{Suppl2}

\paragraph{Model architecture}	 

The model architecture is illustrated in Figure \ref{fig2}, and in more detail in Figure \ref{figKeras}. We designed a CNN made of a sequence of seven convolutional layers followed by a max pooling layer and a fully-connected layer. We used two types of convolutional layers: Regular and Down-Convolution. Regular convolutional layers comprised a 3 x 3 x 3 convolutional operation with 1 x 1 x 1 strides. The down-convolutional layers comprised a 2 x 2 x 2 convolutional operation with 2 x 2 x 2 strides. Each convolutional layer was followed by a Rectified Linear Unit non-linearity \cite{nair2010}. After the last convolutional layer, we used a max pooling layer with a filter size of 2 x 2 x 2. Subsequently, this layer was converted into a fully-connected layer, followed by the output layer containing a single node with a linear activation function. 

\paragraph{Model implementation}	 
	 
The MRI scans were standardized to the spatial resolution of 1 x 1 x 1 millimeters and the size of 256 x 256 x 256 voxels. Additionally, for computational efficiency, they were cropped and down-sampled to 96 x 109 x 96 voxels.

The model was trained on 2100 MRI scans (from 411 subjects) for 200 epochs with the loss function set to the mean absolute error using the Adaptive Moment Estimation optimizer \cite{kingma2014}, a learning rate of $10^{-4}$ and a 3D spatial drop out regularization of 0.9. 

Hyperparameter tuning was carried out on a validation dataset comprising 550 scans of 129 subjects that all had MRI data but did not have genetic data available so that they could not be included in the main analysis. The final evaluation was done on the test set including the 556 subjects of the main analysis that had all MRI, genetic, and gene expression data available. The model performance on the test set is visualized in Figure \ref{fig3}.

\paragraph{Computational comparison with FreeSurfer}	 
	 
Both models where trained on a dual 20 core Intel Xeon 6148 workstation with 768GB RAM equipped with an NVidia Titan-V graphics card.
CNN computations made use of GPU optimization, taking 1 second for the prediction of the volume of the third ventricle per MRI scan. FreeSurfer, which did not utilize the GPU, took 16 hours per MRI scan.

\subsection{Supplementary tables and figures} \label{Suppl3}

\begin{supptable}[!h]
	\centering
	\caption{Descriptive statistics of the $n=556$ individuals in the analyzed sample from the ADNI study. Shown are absolute frequencies for categorical variables, and mean (standard deviation) for quantitative measures.} 
	\label{table2}
        \begin{tabular}{@{}ll@{}}
        \toprule
        \textbf{Measures} & \textbf{Descriptive Statistics}  \\ \midrule
        \textbf{Sample size}	            & 556           \\
        \textbf{Age}, years		            & 72.9 (7.0)    \\
        \textbf{Gender}	                    &  	            \\
        \quad female	                    & 250           \\
        \quad male	                        & 306           \\
        \textbf{Ethnic group}	            &  	            \\
        \quad hispanic/latino	            & 10            \\
        \quad not hispanic/latino	        & 545           \\
        \quad unknown	                    & 1             \\
        \textbf{Education length}, years	& 16.1 (2.8)    \\
        \textbf{APOE $\varepsilon$4 allele}	&  	            \\
        \quad homozygot minor allele	    & 331           \\
        \quad heterozygot	                & 186           \\
        \quad homozygot major allele	    & 39            \\
        \textbf{Cognitive status}	        &  	            \\
        \quad cognitive normal	            & 194           \\
        \quad mild cognitive impairment	    & 338           \\
        \quad dementia					    & 24            \\ \bottomrule
        \end{tabular}%
\end{supptable}

\begin{suppfigure}
	\centering
	\hspace*{-0.275in}
	\includegraphics[width=1.1\linewidth]{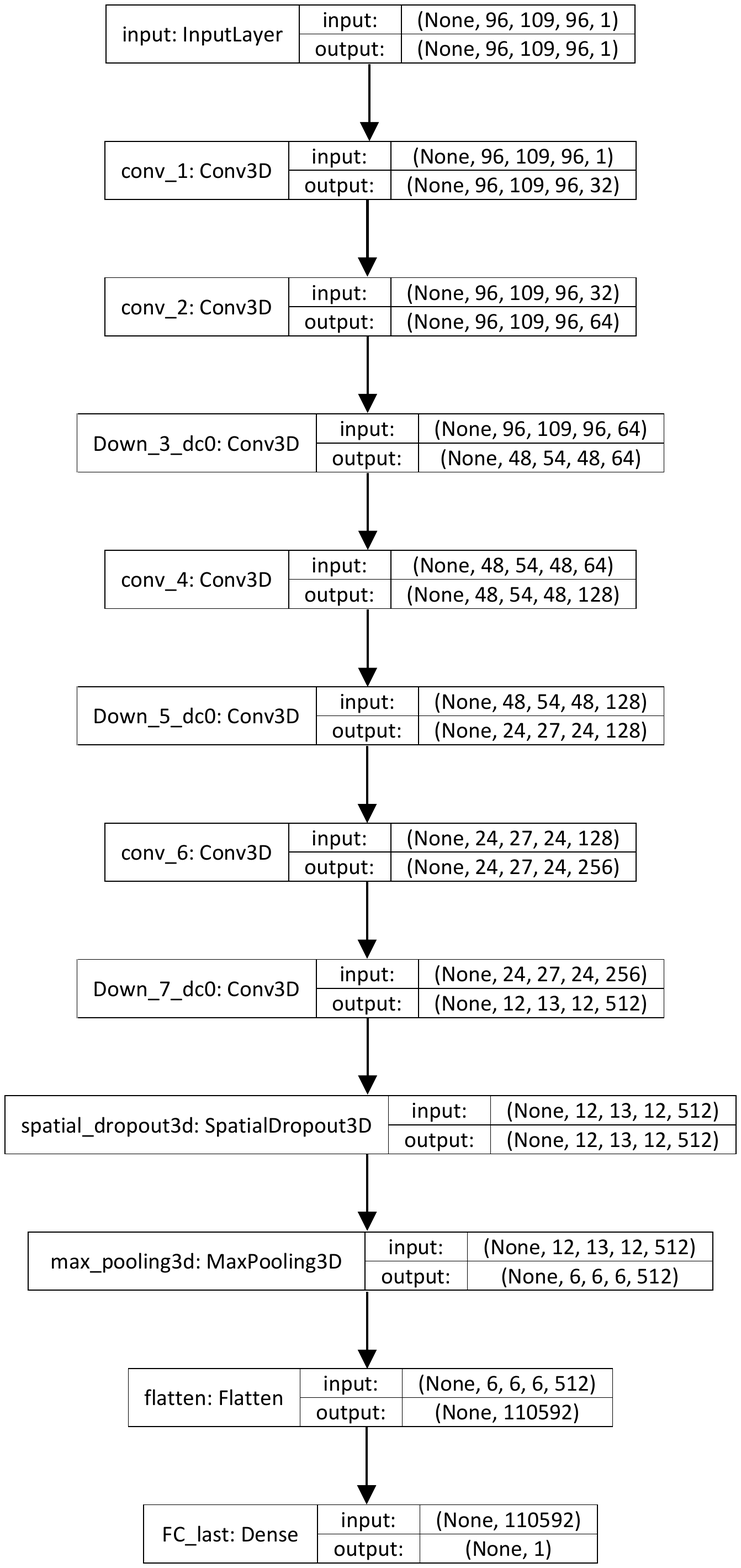}
	\caption{Graphical visualization of the 3D convolutional neural network model in Keras. Shown are input and output of the different layers, and the respective voxels and channels. For example, the input volume had 96 $\times$ 109 $\times$ 96 voxels and 1 channel. As all computations were done in one batch, the batch size was not specified (noted as "None" in the graph).}
	\label{figKeras}
\end{suppfigure}

\begin{suppfigure}
	\centering
	\includegraphics[width=0.9\linewidth]{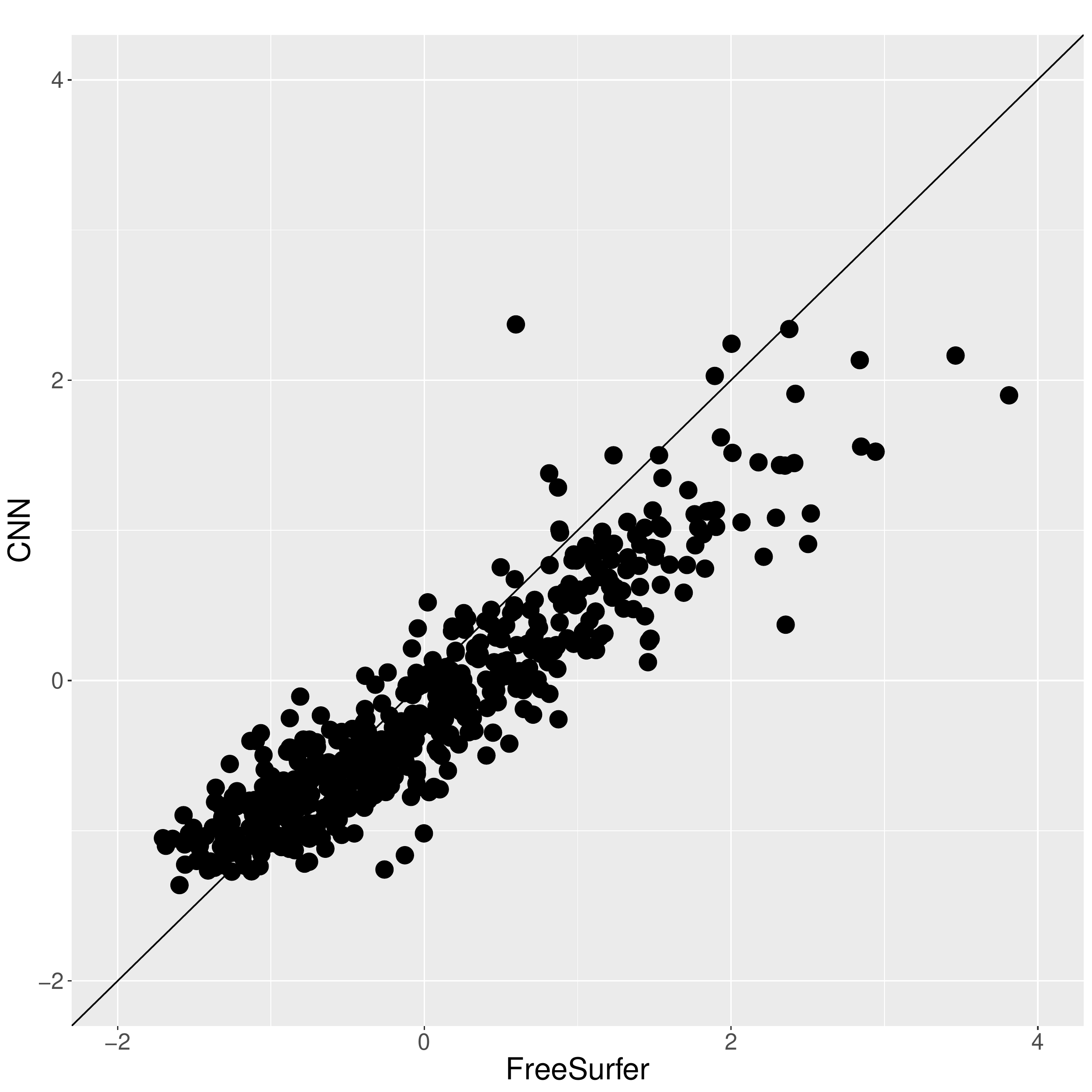}
	\caption{Scatterplot of the volume of the third ventricle prediction by FreeSurfer (x axis) and the CNN (y axis). All predictions are represented as $z$ scores. In addition, the diagonal is printed for a comparison of both predictions.} 
	\label{fig3}
\end{suppfigure}

\begin{suppfigure}[!b]
	\centering
	\includegraphics[width=0.9\linewidth]{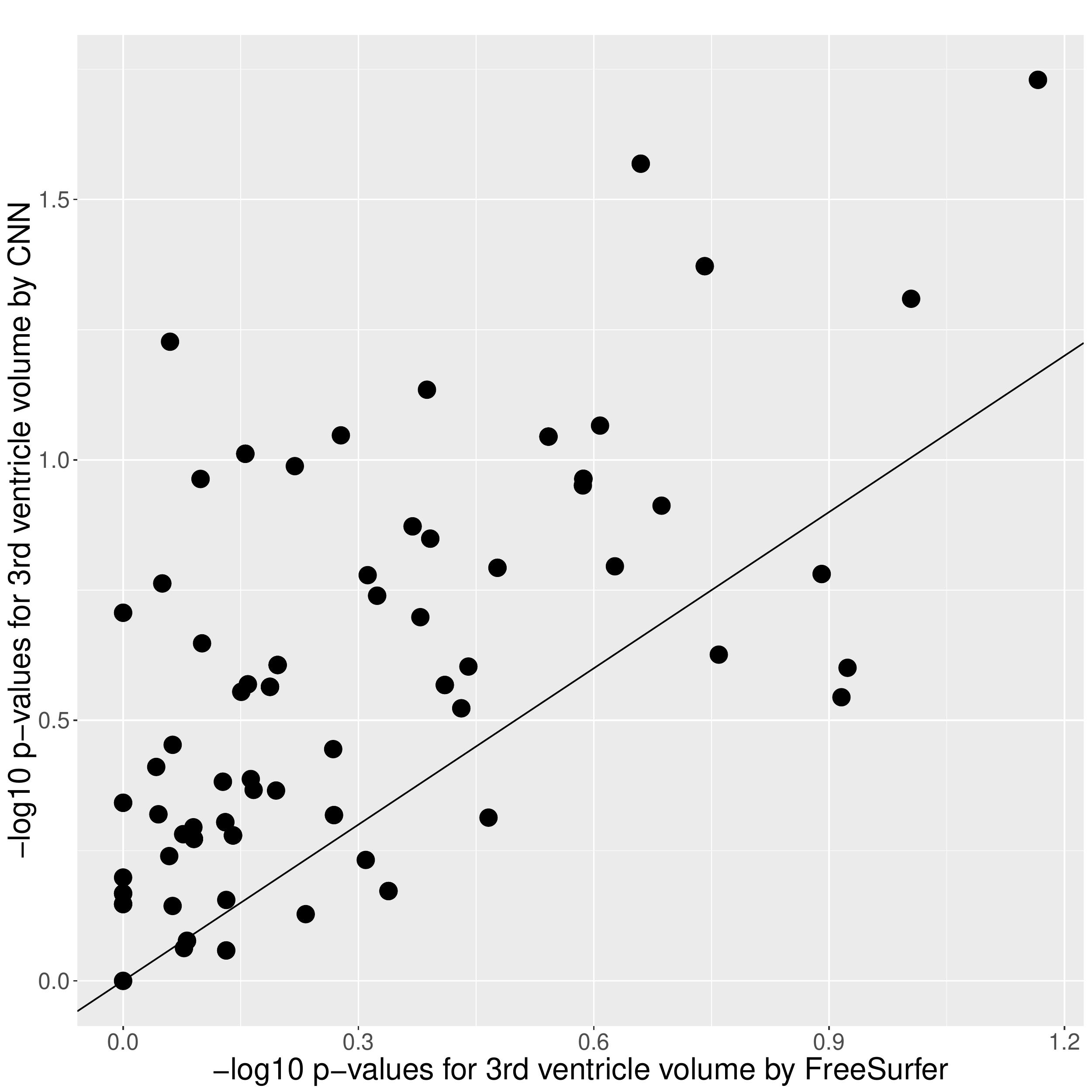}
	\caption{Scatterplot of the -log$_{10}$ p-values from association tests of the 125 genes with the volume of the third ventricle predicted by FreeSurfer (x axis) and CNN (y axis) as outcome, using SKAT-O for testing. In addition, the diagonal is printed for a comparison of both tests.}
	\label{fig4}
\end{suppfigure}

\begin{suppfigure}[!b]
	\centering
	\includegraphics[width=0.9\linewidth]{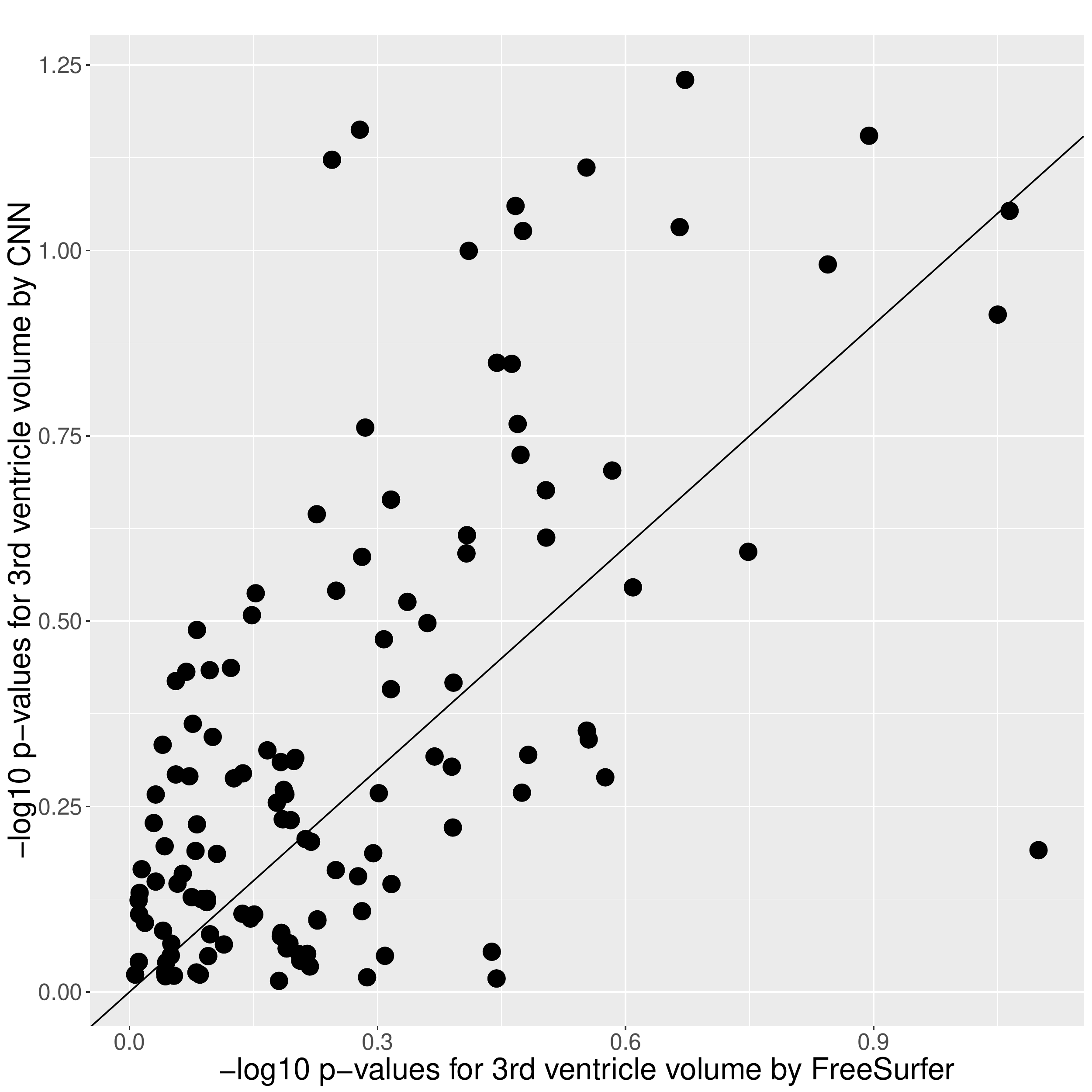}
	\caption{Scatterplot of the -log$_{10}$ p-values from association tests of the 125 genes with the volume of the third ventricle predicted by FreeSurfer (x axis) and CNN (y axis), using the new kernel-based test 1 (identity $V$, MAF + omics $W$). In addition, the diagonal is printed for a comparison of both tests.}
	\label{fig5}
\end{suppfigure}

\begin{suppfigure}[!b]
	\centering
	\hspace*{-0.7in}
	\includegraphics[width=1.2\linewidth]{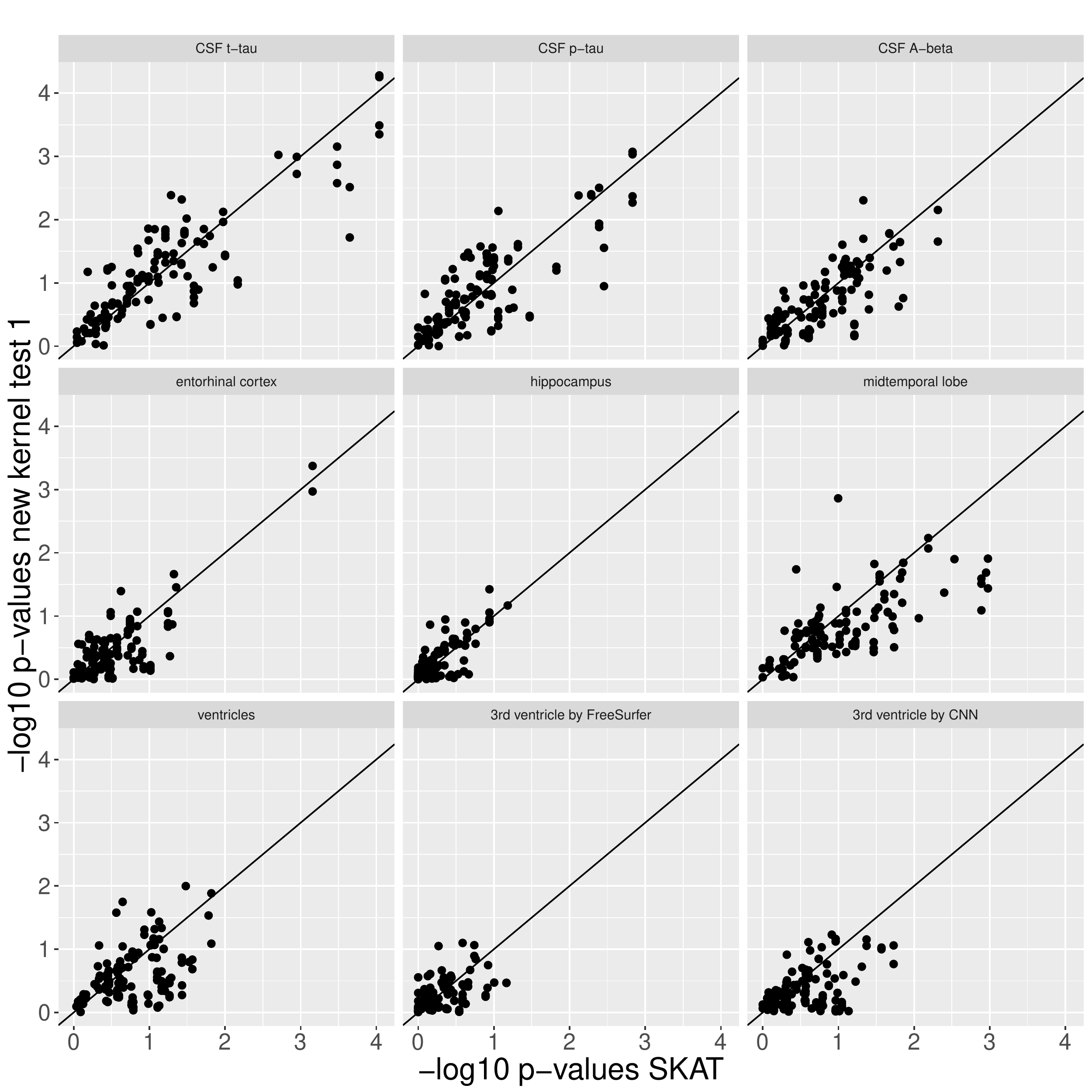}
	\caption{Scatterplot of the -log$_{10}$ p-values from association tests of the 125 genes using SKAT (x axis) and the new kernel-based test 1 (identity $V$, MAF + omics $W$, y axis), for each of the 9 traits in separate panels. In addition, the diagonal is printed for a comparison of both tests.}
	\label{fig6}
\end{suppfigure}

\end{document}